\documentclass[10pt,twocolumn,letterpaper]{article}

\usepackage{cvpr}
\usepackage{times}
\usepackage{epsfig}
\usepackage{graphicx}
\usepackage{amsfonts}
\usepackage{amsmath}
\usepackage{amssymb}

\usepackage{nicefrac}       
\usepackage{microtype}      
\usepackage{wrapfig}

\usepackage{multirow}
\usepackage{verbatim}
\usepackage{color}
\usepackage{float}
\usepackage{enumitem}
\usepackage{booktabs}
\usepackage{tabulary,multirow,overpic,xcolor}
\usepackage[caption=false]{subfig}

\usepackage[british,UKenglish,USenglish,english,american]{babel}

\definecolor{citecolor}{RGB}{34,139,34}
\usepackage[pagebackref=true,breaklinks=true,letterpaper=true,colorlinks,
  citecolor=citecolor,bookmarks=false]{hyperref}

\cvprfinalcopy 

\def\cvprPaperID{945} 

\begin{document}

\title{Binge Watching: Scaling Affordance Learning from Sitcoms}

\author{Xiaolong Wang\thanks{Indicates equal contribution.} 
\quad     \quad  Rohit Girdhar\footnotemark[1]  \quad  \quad  Abhinav Gupta  \\
The Robotics Institute, Carnegie Mellon University \\
{\tt\small \url{http://www.cs.cmu.edu/~xiaolonw/affordance.html}}
}

\maketitle
\begin{abstract}
In recent years, there has been a renewed interest in jointly modeling perception and action. At the core of this investigation is the idea of modeling affordances\footnote{Affordances are opportunities of interaction in the scene. In other words, it represents what actions can the object be used for.}. However, when it comes to predicting affordances, even the state of the art approaches still do not use any ConvNets. Why is that? Unlike semantic or 3D tasks, there still does not exist any large-scale dataset for affordances. In this paper, we tackle the challenge of creating one of the biggest dataset for learning affordances. We use seven sitcoms to extract a diverse set of scenes and how actors interact with different objects in the scenes. Our dataset consists of more than 10K scenes and 28K ways humans can interact with these 10K images. We also propose a two-step approach to predict affordances in a new scene. In the first step, given a location in the scene we classify which of the 30 pose classes is the likely affordance pose. Given the pose class and the scene, we then use a Variational Autoencoder (VAE)~\cite{Kingma14} to extract the scale and deformation of the pose. The VAE allows us to sample the distribution of possible poses at test time. Finally, we show the importance of large-scale data in learning a generalizable and robust model of affordances.
\end{abstract}

\section{Introduction}
One of the long-term goals of computer vision, as it integrates with robotics, is to translate perception into action. While 
vision tasks such as semantic or 3D understanding have seen remarkable improvements in performance, the 
task
of translating perception into actions has not seen any major gains. For example, the state of the art approaches in predicting affordances still do not use any ConvNets with the exception of ~\cite{Fouhey15b}. Why is that? What is common across the tasks affected by ConvNets is the availability of large scale supervisions. For example, in semantic tasks, the supervision comes from crowd-sourcing tools like Amazon Mechanical Turk; and in 3D tasks, supervision comes from structured light cameras such as the Kinect. But no such datasets 
exist for supervising actions afforded by a scene. 
Can we create a large-scale dataset that can alter the course in this field as well?

\begin{figure}
    \centering
    \includegraphics[width=0.5\textwidth]{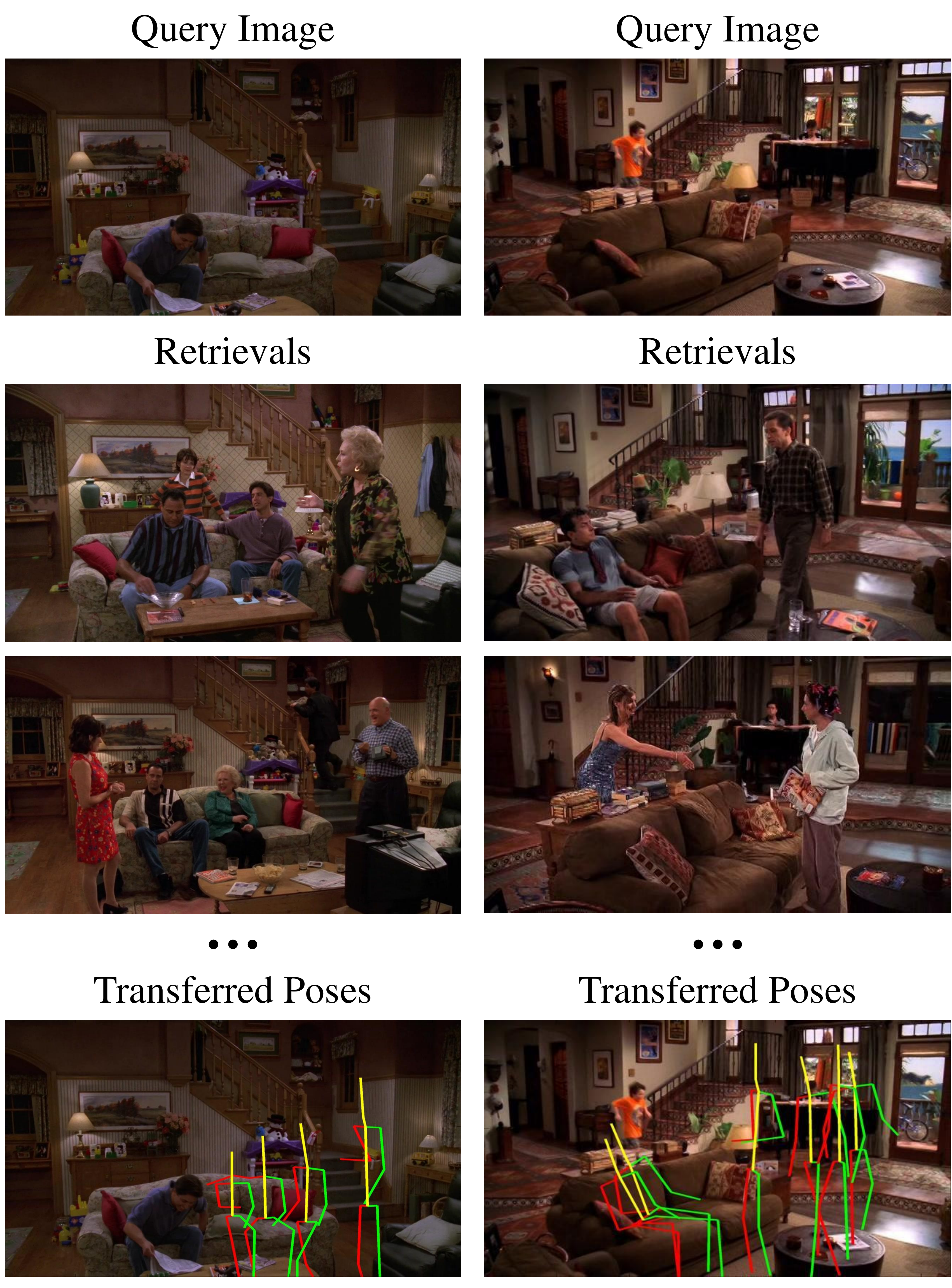}
    \caption{We  propose  to  binge-watch  sitcoms  to  extract  one  of  the largest affordance datasets ever. We use more than 100M frames from seven different sitcoms to find empty scenes and same scene with humans. This allows us to create a large-scale dataset with scenes and their affordances.}
    
    \label{fig:teaser}
\end{figure}

There are several possible ways to create a large-scale dataset for affordances: 
(a) First option is to label the data: given empty images of room, we can ask mechanical turkers to label what actions can be done at different locations. However, labeling images with affordances is extremely difficult and an unscalable solution. (b) The second option is to automatically generate data by doing actions themselves. One can either use robots and reinforcement learning to explore the world and the affordances. However, collecting large-scale diverse data is not yet scalable in this manner. (c) A third option is to use simulation: one such example is ~\cite{Fouhey15b} where they use the block geometric model of the world to know where 
human skeletons would fit.
However, this model only captures physically likely actions and does not capture the statistical probabilities behind every action. For example, it allows predictions such as humans can sit on top of stoves; and for the open space near doors it predicts walking as the top prediction (even though it should be reaching the door).

In this paper, we propose another alternative: watch the humans doing the actions and use those to learn affordances of objects. But how do we find large-scale data to do that? We propose to binge-watch sitcoms to extract one of the largest affordance datasets ever. Specifically, we use every episode and every frame of seven sitcoms~\footnote{How I Met Your Mother, Friends, Two and a Half Men, Frasier, Seinfield, The Big Bang Theory, Everybody Loves Raymond} which amounts to processing more than 100 Million frames to extract parts of scenes with and without humans. We then perform automatic registration techniques followed by manual cleaning to transfer poses from scenes with humans to scenes without humans. This leads to a dataset of 28882 poses in empty scenes. 

We then use this data to learn a mapping from scenes to affordances. Specifically, we propose a two-step approach. In the first step, given a location in the scene we classify which of the 30 pose classes (learned from training data) is the likely affordance pose. Given the pose class and the scene, we then use the Variational Autoencoder (VAE) to extract the scale and deformation of the pose. Instead of giving a single answer or averaging the deformations, VAE allows us to sample the distribution of possible poses at test time. We show that training an affordance model on large-scale dataset leads to a more generalizable and robust model.

\section{Related Work}
The idea of affordances~\cite{Gibson79} was proposed by James J. Gibson  in late seventies, where 
he described
affordances 
as ``opportunities for interactions'' provided by the environment. Inspired by Gibson's ideas, our field has time and again fiddled with the idea of functional recognition~\cite{Stark, Rivlin}. In most cases, the common approach is to first estimate physical attributes and then reason about affordances. Specifically, manually-defined rules are used to reason about shape and geometry to predict affordances~\cite{Stark, Binford}. However, over years, the idea of functional recognition took backstage because these approaches lacked the ability to learn from data and handle the noisy input images. 

On the other hand, we have made substantial progress in the field of semantic image understanding. This is primarily due to the result of availability of large scale training datasets~\cite{Deng09,COCO} and high capacity models like ConvNets~\cite{lecun90, krizhevsky_nips12_alexnet}. However, the success of ConvNets has not resulted in significant gains for the field of functional recognition. Our hypothesis is that this is due to the lack of large scale training datasets for affordances. While it is easy to label objects and scenes, labeling affordances is still manually intensive. 

There are two alternatives to overcome this problem. First is to estimate affordances by using reasoning on top of 
semantic~\cite{Dai16,He2015,deeplab}
and 3D~\cite{Eigen15,Bansal16,3dinterpreter,Girdhar16b}
scene understanding. There has been a lot of recent work which follow  this alternative: \cite{Gupta07,Kjellstrom08} model relationship between semantic object classes and actions; Yao et al.~\cite{Yao10a} model relationships between object and poses. These relationships can be learned from videos~\cite{Gupta07}, static images~\cite{Yao10a} or even time-lapse videos~\cite{Delaitre12}. Recently, \cite{Zhu14} 
proposed a way to reason about object affordances by combining object
categories and attributes in a knowledge base manner. Apart from using semantics, 3D properties have also been used to estimate affordances ~\cite{Gupta11, Grabner11, Fouhey12, Zhao13, Xie13}. Finally, there have been efforts to use specialized sensors such as Kinect to estimate geometry followed by estimating affordances as well ~\cite{Jiang13,Koppula14,Koppula15,Zhu15}.

While the first alternative tries to estimate affordances in low-data regime, a second alternative is to collect data for affordances without asking humans to label each and every pixel. One possible way is to have robots themselves explore the world and collect data for how different objects can used. For example, ~\cite{pintoicra16} uses self-supervised learning to learn grasping affordances of objects or \cite{Pulkit_16,pintoeccv2016} focus on learning pushing affordances. However, using robots for affordance supervision is still not a scalable solution since collecting this data requires a lot of effort. Another possibility is to use simulations~\cite{mottaghiECCV16}. For example, Fouhey et al.~\cite{Fouhey15b} propose a 3D-Human pose simulator which can collect large scale data using 3D pose fitting. But this data only captures physical notion of affordances and does not capture the statistical probabilities behind every action. In this work, we propose to collect one of the biggest affordance datasets using sitcoms and minimal human inputs. Our approach sifts through more than 100M frames to find high-quality scenes and corresponding human poses to learn affordance properties of objects.

\begin{figure*}
    \centering
    \includegraphics[width=1\textwidth]{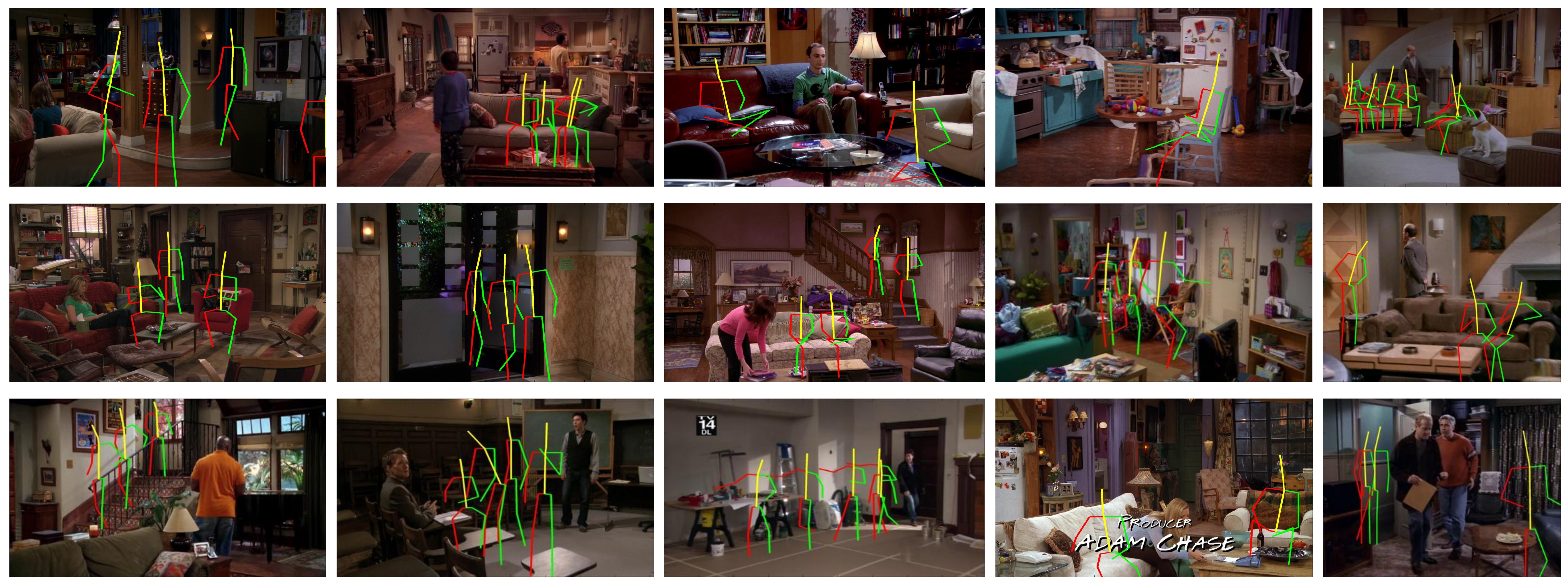}
    \caption{Some example images from Sitcom Affordance dataset. Note that our images are quite diverse and we have large samples of possible actions per image.}
    \label{fig:examples}
\end{figure*}

\section{Sitcom Affordance Dataset}
Our first goal towards data-driven affordances is to collect a  large scale dataset for affordances. What we need is an image dataset of scenes such as living rooms, bedrooms etc and what actions can be performed in different parts of the scene. In this paper, inspired by some recent work~\cite{gupta_cvpr11}, we represent the output space of affordances in terms of human poses. But where can we find images of the same scene with or without people in it?

The answer to the above question lies in exploiting the TV Sitcoms. In sitcoms, characters share a common environment, such as a home or workplace. A scene with exact configuration of objects appears again and again as multiple episodes are shot in it. For example, the living room in Friends appears in all the 10 seasons and ~240 episodes and each actor labels the action space in the scene one by one as they perform different activities. 

We use seven such sitcoms and process more than 100M frames of video to create the largest affordance dataset. We follow a three-step approach to create the dataset: (1) As a first step, we mine the 100M frames to find empty scenes or sub-scenes. We use an empty scene classifier in conjunction with face and person detector to find these scenes; (2) In the second step, we use the empty scenes to find the same scenes but with people performing actions. We use two strategies to search for frames with people performing actions and transfer the estimated poses~\cite{carreira_arxiv2015_ief}  to empty scenes by simple alignment procedure; (3) In the final step, we perform manual filtering and cleaning to create the dataset. We now describe each of these steps in detail.

\begin{figure*}
    \centering
    \includegraphics[width=1\textwidth]{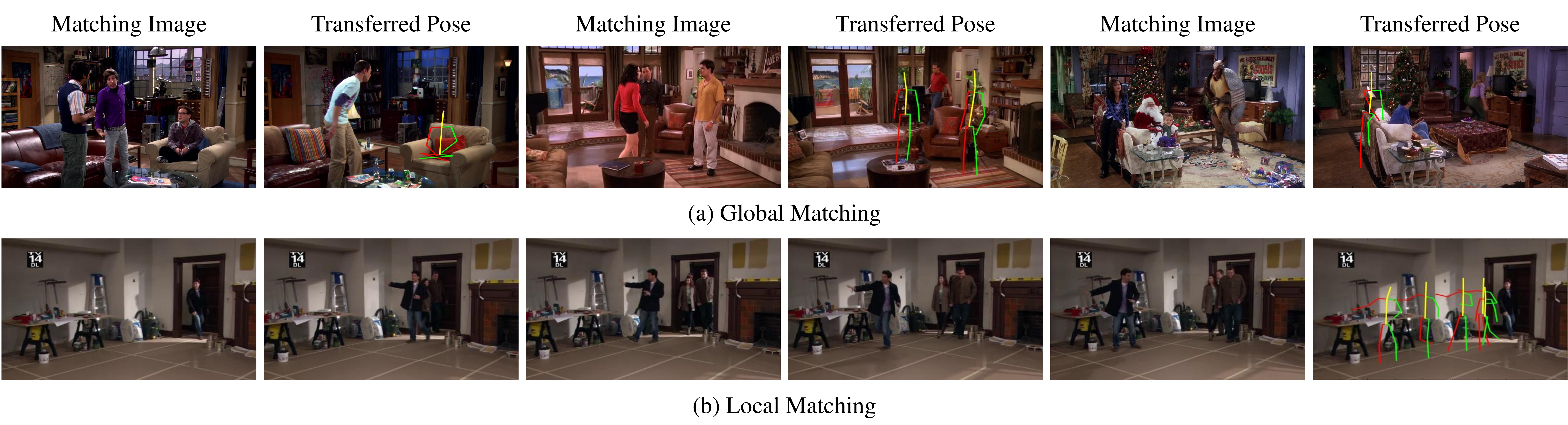}
    \caption{We propose to use two approaches to transfer poses. In global matching approach, we match an empty scene to all the images in the sitcom. Sometimes the matches occur in different seasons. Given these matches, we transfer poses to the image. In the local matching approach, we use the next 5-10 sec of video to transfer poses via optical-flow registration scheme.}
    \label{fig:dataset}
\end{figure*}

\subsubsection*{Extracting Empty Scenes}
We use a combination of three different models to extract empty scenes from 100M frames: face detection, human detection and scene classification scores. In our experiment, we find face detection~\cite{liang_arxiv15} is the most reliable criteria. Thus, we first filter out scenes based on the size of the largest face detected in the scenes. We also applied Fast-RCNN to detect humans~\cite{girshickICCV15fastrcnn} in the scene. We reject the scenes where humans are detected. Finally, we have also trained a CNN classifier for empty scenes. The positive training data for this classifier are scenes in SUN-RGBD~\cite{song_cvpr15_sunrgbd} and  MIT67~\cite{quattoni_cvpr09_mit67}; the negative data are random video frames from the TV series and Images-of-Groups~\cite{gallagher_cvpr09_group}. The classifier is finetuned on PlaceNet~\cite{zhou_nips14_placenet}. After training this classifier, we apply it back on the TV series training data and select 1000 samples with the highest prediction scores. We manually label these 1000 images and use them to  fine-tuned the classifier again. This ``hard negative'' mining procedure turns out to be very effective and improve the generalization power of the CNN across all TV series.

\subsubsection*{People Watching: Finding Scenes with People}

We use two search strategies to find scenes with people. Our first strategy is to use image retrieval where we use empty scenes as query images and all the frames in the TV-series as retrieval dataset. We use cosine distance on the pool5 features extracted by ImageNet pre-trained AlexNet. In our experiments, we find the pool5 features are robust to small changes of the image, such as the decorations and number of people in the room, while still be able to capture the spatial information. This allows us to directly transfer 
human skeletons from matching images to the query image. We show some examples of data generated using this approach in the top two rows of Fig.~\ref{fig:dataset}. 

Besides global matching of frames across different episodes of TV shows, we also transfer human poses within local shots (short clips at most 10 seconds) of videos. Specifically, given one empty frame we look into the video frames ranging from 5 seconds before this frame to 5 seconds after it. We  perform pose estimation on every frame. We then compute the camera motions of each frame with respect to the empty frame by accumulating the optical flows. Given these motion information, we can map the detected poses to the empty frame, as shown in the bottom two rows in Fig.~\ref{fig:dataset}.

\subsubsection*{Manual Annotations}

\begin{wrapfigure}{r}{0.2\textwidth}

\label{tab:dataStats}
\begin{tabular}{@{}rc@{}}
\toprule
{\footnotesize Source}              &  {\footnotesize \#Datapoints}\\ \midrule
{\footnotesize HIMYM}      & {\footnotesize 3506}  \\
{\footnotesize TBBT}       & {\footnotesize 3997} \\
{\footnotesize Friends}    & {\footnotesize 3872}  \\ 
{\footnotesize TAAHM}      & {\footnotesize 3212}  \\
{\footnotesize ELR}        & {\footnotesize 5210} \\
{\footnotesize Frasier}    & {\footnotesize 6018}  \\ 
{\footnotesize Seinfeld}   & {\footnotesize 3067}  \\ 
{\footnotesize Total}      & {\footnotesize 28882}  \\ 
\bottomrule
\end{tabular}
\setlength{\columnsep}{10pt}
\end{wrapfigure}

Our goal is to use the automated procedure above to generate valid hypothesis of possible poses in empty scenes. However, the alignment procedure is not perfect by any means. Thus, we also need human annotators to manually adjust pose joints by scaling and translating. In this way, the pose in the empty scene can be aligned with the human in the image where the pose is transferred from. In cases where the poses are not fitting with the scene, due to occlusions or incorrect matching, we remove such poses. The final dataset we obtain contains 28882 human poses inside 11449 indoor scenes. The detailed statistics of how many poses for each TV series are summarized in Table~\ref{tab:dataStats}.

\section{VAEs for Estimating Affordances}
Given an indoor scene and the location, we want to predict what is the most likely human pose. One naive approach is training a ConvNet with the image and the location as input, predict the heat maps for each joint of the pose as state-of-the-art pose estimation approaches~\cite{carreira_arxiv2015_ief}. However, our problem is very different from standard pose estimation, since we do not have the actual human which can provide the pose structure and regularize the output. 

We explore an alternative way: we decompose the process of predicting poses into two stages: (i) categorical prediction: we first cluster all the human poses in the dataset into 30 clusters, and predict which pose cluster is most likely given the location in the scene; (ii) given the pose cluster center, we predict its scale as well as the deformations for pose joints such that it fits into the real scene.

\begin{figure*}[t]
\center
\includegraphics[width=\textwidth]{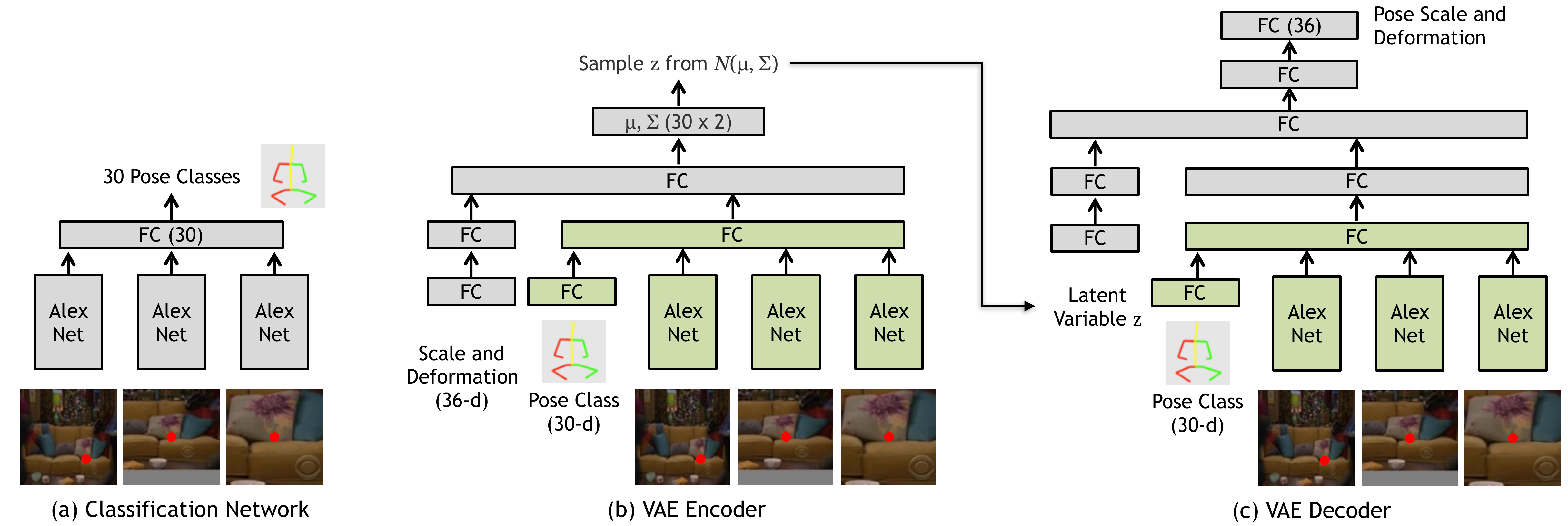}
\caption{\label{fig:models}
Our Affordance Prediction Model. The encoder and decoder in VAE share the weights which are highlighted as green. All fully connected  layers have 512 neurons unless it is specified in the figure.}
\end{figure*}

\subsection{Pose Classification}
\begin{figure}[t]
\center
\includegraphics[width=0.45\textwidth]{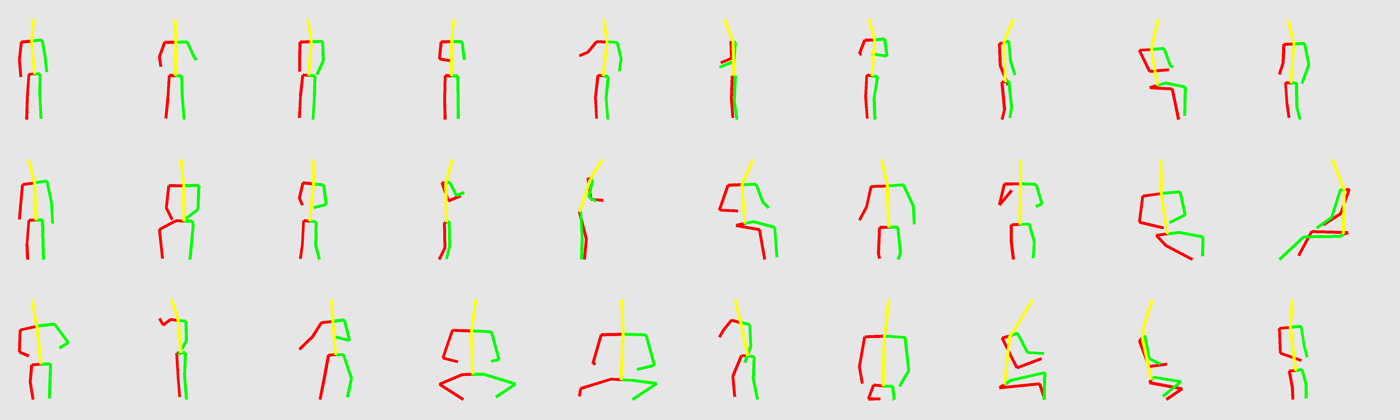}
\caption{\label{fig:clusters}
Cluster centers of human poses in sitcom dataset. These clusters are used as pose categories predicted by classification network.}
\end{figure}

As a first step, given an input image and a location, we first do a categorical prediction of human poses. But what are the right categories? We use a data-driven vocabulary in our case. Specifically, we use randomly sampled 10K poses from the training videos. We then compute the distances between each pair of poses using procrustes analysis
over the 2D joint coordinates, and cluster them into 30
clusters using k-mediod clustering. We visualize the centers of the clusters as Fig.~\ref{fig:clusters}. 

In the first stage of prediction, we train a ConvNet which uses the location and the scene as input and predict the likely pose class. Note that multiple pose classes could be reasonable in a particular location (e.g. one can stand before the chair or sit on the chair), thus we are not trying to regress the exact pose class. Instead we predict a probability distribution over all classes and select the most likely one. The selected pose center can be further adjusted to fit in the scene in the second stage. 

\noindent {\bf Technical Details:} The input to the ConvNet is the image and the location where to predict likely pose. To represent this point in the image, we crop a square patch using it as the center and the side length is the height of the image frame. We also crop another patch in a similar way except that the length is half the height of the image. As illustrates in Fig.~\ref{fig:models} (a), the red dots on the input images represent the location. The two cropped patches can offer different scales of information and we also take the whole image as input. The 3 input images are all re-scaled to $227 \times 227$. 

Given the 3 input images, they are forwarded to 3 ConvNets which share the weights between them. We apply the AlexNet architecture~\cite{krizhevsky_nips12_alexnet} for the ConvNet here and concatenate the 3 $fc7$ outputs. The concatenated feature is further fully connected to 30 outputs, which represents 30 pose classes. During training, we apply SoftMax classification loss and the AlexNet is pre-trained with ImageNet dataset~\cite{deng_cvpr09_imagenet}. 

\subsection{Scale and Deformation Estimation}
Given the pose class and scene, we need to predict the scale and the deformations of each joint to fit the pose into the scene. However, the scale and deformations of the pose are not deterministic and there could be ambiguities. For example, given an empty floor and a standing pose class, it could be a child standing there (which corresponds to a smaller scale) or an adult standing there (which corresponds to a larger scale). Thus instead of directly training a ConvNet to regress the scale  and deformations, we apply the conditional Variational Auto-Encoder (VAE)~\cite{Kingma14,vae_eccv2016} to generate the scale and deformations conditioned on the input scene and pose class. 

\textbf{Formulations for the conditional VAE.} We applied the conditional VAE to model the scale and deformations of the estimated pose class. For each sample, we define the deformations and scale as $y$, the conditioned input images and pose class as $x$, and the latent variables sampled from a distribution $Q$ as $z$. Then the standard variational equality can be represented as, 
\begin{flalign}\label{eq:vaeeq}
& \log P(y|x) - KL[Q(z|x,y) || P(z|x,y)] \nonumber \\ 
& = E_{z\sim Q}[\log P(y|z,x)] - KL[Q(z|x,y) || P(z|x)], 
\end{flalign}
where $KL$ represents the KL-divergence between the distribution $Q(z|x,y)$ and the distribution $P(z|x)$. Note that in VAE, we assume $P(z|x)$ is a normal distribution $\mathcal{N}(0,1)$. The distribution $Q$ is another normal distribution which can be represented as $Q(z|x,y) = Q(z|\mu(x,y),\sigma(x,y))$, where $\mu(x,y)$ and $\sigma(x,y)$ are estimated via the encoder in VAE. The log-likelihood $\log P(y|z,x)$ is modeled by the decoder in VAE. The details are explained as below. 

\textbf{Encoder.} As Fig.~\ref{fig:models}(b) illustrates, the inputs of model include 3 images which are obtained in the same way as the classification network, a 30-d binary vector indicating the pose class (only one dimension is activated as 1), and a 36-d vector representing the ground truth scale and deformations. The images are fed into the AlexNet model and we extract the $fc7$ feature for each of them. The pose binary vector and the vector of scale and deformations are both forwarded though two fully connected layers. Each fully connected layer has 512 neurons. The outputs for each components are then concatenated together and fully connected to the outputs. The dimension of the outputs is $30 \times 2$ which are two vectors of mean $\mu(x,y)$ and variance $\sigma(x,y)$ of the distribution $Q$.

We calculate the ground truth scale for the height of pose $s_h$ as the actual pose height divided by the normalized height (ranging from 0 to 1) of cluster center. The ground truth scale for the width $s_w$ is calculated in a similar way. Given the ground truth $(s_h, s_w)$, we can re-scale the cluster center and aligned it to the input location. The deformation for each joint is the spatial distance $(dx, dy)$ between  the scaled center and ground truth pose. Since we have 17 pose joints $(dX, dY) = (dx_1, dy_1,...,dx_{17},dy_{17})$, there are 34 outputs representing the deformations. Together with the scale $s_h, s_w$, the outputs of the generator are 36 real numbers.

\textbf{Decoder.} As Fig.~\ref{fig:models}(c) shows, the decoder has a similar architecture as the encoder. Instead of taking a vector of scale and deformations as input, we replace it with the latent variables $z$. The output of the network is changed to a 36-d vector of scale and deformations whose groundtruth is identical to the 36-d input vector of the encoder. Note that we share the feature representations for the conditional variables (images and classes) between the encoder and decoder. 

\textbf{Training.} As indicated by Eq.~\ref{eq:vaeeq}, we have two losses during training the VAE model. To maximize the log-likelihood, we apply a Euclidean distance loss to minimize distance between the estimated scale and deformations $y^{*}$ and the ground truths as, 
\begin{flalign}\label{eq:l1}
& L_{1} = || y^{*} - y ||^2.
\end{flalign}
And the other loss is to minimize the KL-divergence between the estimated distribution $Q$ and the normal distribution $\mathcal{N}(0,1)$ as,
\begin{flalign}\label{eq:l2}
& L_{2} = KL[Q(z|\mu(x,y),\sigma(x,y)) || \mathcal{N}(0,1)]. 
\end{flalign}

Note that the first loss $L_1$ is applied on top of the decoder model, and its gradient is backpropagated though all the layers. To do this, we follow the reparameterization trick introduced in~\cite{VAE14}: we represent the latent variables as $z=\mu(x,y) + \sigma(x,y) \cdot \alpha$, where $\alpha$ is a variable sampled from $\mathcal{N}(0,1)$. In this way, the latent variables $z$ is differentiable with respect to $\mu$ and $\sigma$.

\subsection{Inference} \label{sec:inference}

Given the trained models, we want to tackle two tasks: (i) generating poses on an empty location of a scene and (ii) estimate if a pose fits the scene or not. 

For the first task, given an image and a point representing the location in the image, we first perform pose classification and obtain the normalized center pose of the corresponding class. The classification scores, together with the latent variables $z$ sampled from $\mathcal{N}(0,1)$ and images are forwarded to the VAE decoder model (Fig.~\ref{fig:models} (c)). We scale the normalized center with the inferred scale $(s_h^{*}, s_w^{*})$ and align the pose with the input point. Then we adjust each joint of the pose by adding the deformations $(dX^{*},dY^{*})$. 

For the second task, we want to estimate whether a given pose fits the scene or not. To do this, we first perform the same estimation of the pose given an empty scene as the first task, then we compute the euclidean distance $D$ between the estimated pose and the given pose. To ensure the robustness of the estimation, we repeat this procedure by sampling different $z$ for $m=10$ times, and calculate the average distance $\frac{1}{m} \sum_{1}^{m} D_i$ as the final result. If the final distance is less than a threshold $\delta$, then the given pose is taken as a reasonable pose.

\begin{figure*}[t]
\center
\includegraphics[width=\textwidth]{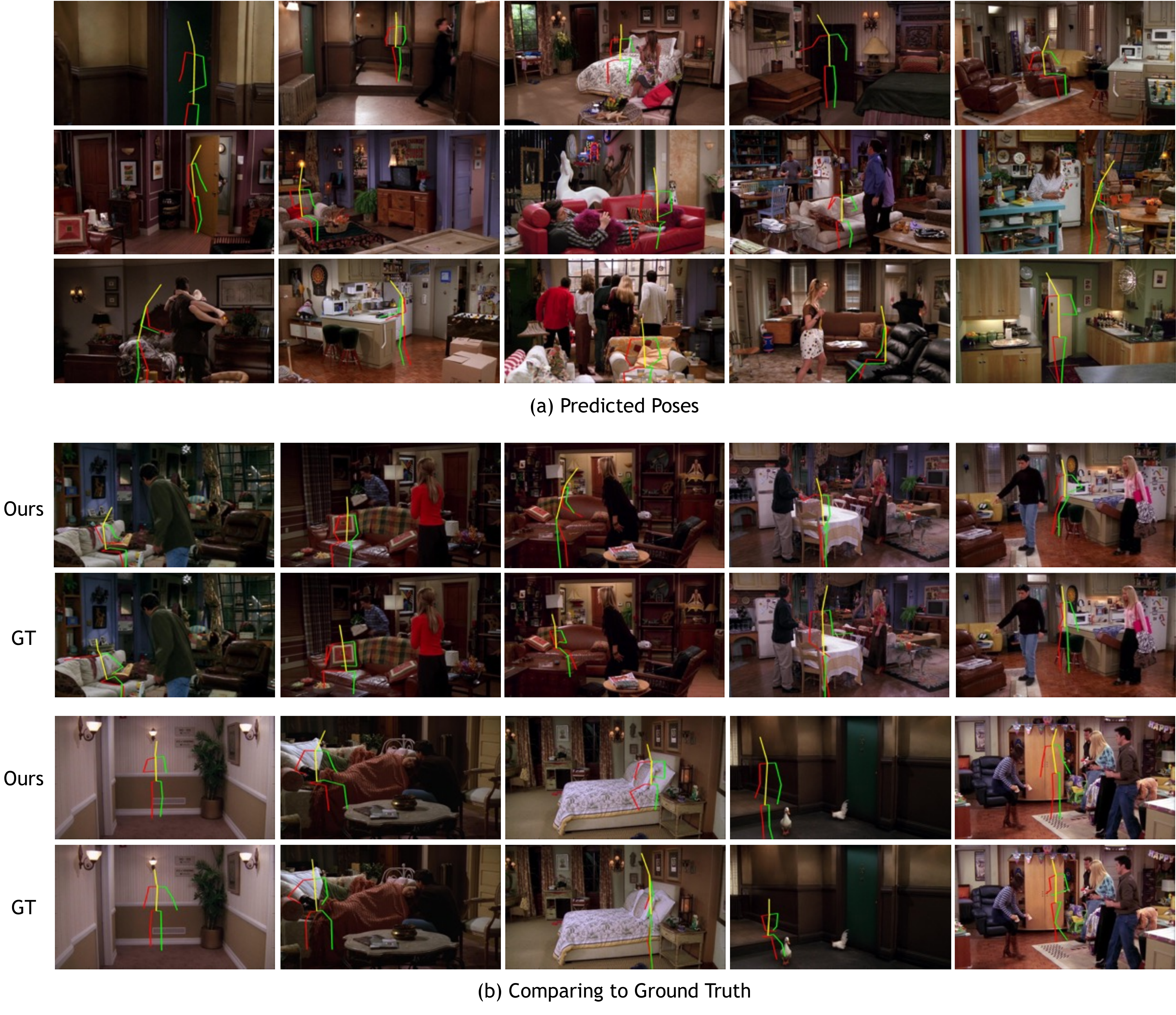}
\vspace{-0.2in}
\caption{\label{fig:results}
Qualitative results of generating poses: We show qualitative results on scenes from Friends dataset. (a) As we can see the human poses generated seem very reasonable. (b) We also compare the poses generated using our approach with the ground truth poses.}
\vspace{-0.05in}
\end{figure*}

\section{Experiments and Results}
We are going to evaluate our approach on two tasks: (i) affordance prediction: given an input image and a location, generate the likely human pose at that location; (ii) classify whether a given pose in a scene is possible or not. 

We train our models using data collected from the TV series of ``How I Met Your Mother'', ``The Big Bang Theory'', ``Two and A Half Man'', ``Everyone Loves Raymond'', ``Frasier'' and ``Seinfeld''. The models are tested on the frames collected from ``Friends''. For training data, we have manually filtered and labeled 25010 accurate poses over 10009 different scenes. For testing data, we have collected 3872 accurate poses over 1490 different scenes and we have also artificially generated 9572 poses in the same scenes which are either physically impossible or very unlikely to happen in the real world. 

During training, we initialize the AlexNet image feature extractor with ImageNet pre-training and the other layers are initialized randomly.  We apply the Adam optimizer during training with learning rate $0.0002$ and momentum term $\beta_1 = 0.5, \beta_2 = 0.999$. To show that large scale of data matters, we perform the experiments on different size of the dataset. 

We also evaluate the performance of our approach as the training dataset size increases. Specifically, we randomly sample 2.5K and 10K of data for training, and compare these models with our full model which uses 25K data for training. 

\paragraph{Baseline}
We compare our VAE approach to a heatmap regression based baseline.
Essentially, we represent the human skeletons as a 17-channel heatmap, one for each joint. We train a three-tower AlexNet (upto conv5) architecture,
where each tower looks at a different scale of the image around the given point. The towers have shared parameters and are initialized with ImageNet. The outputs are concatenated across the towers and passed through a convolution and deconvolution layer to produce a 17 channel heatmap, which is trained with euclidean loss.

\subsection{Generating poses in the scenes}
As we mentioned in the approach, we generate the human pose via estimating the pose class and the scale as well as deformations. 

\noindent \textbf{Qualitative results.} We show our prediction results as Fig.~\ref{fig:results}(a). We have shown that our model can generate very reasonable poses including sitting on a coach, closing the door and reaching to the table, etc. We also compare our results with the ground truth as Fig.~\ref{fig:results}(b). We show that we can generate reasonable results even though it can be very different from the ground truth poses. For example, in the 3rd column of the 2nd row, we predict a pose sitting on a bed while the ground truth is standing in front of the bed. 

\begin{figure*}[t]
\center
\includegraphics[width=\textwidth]{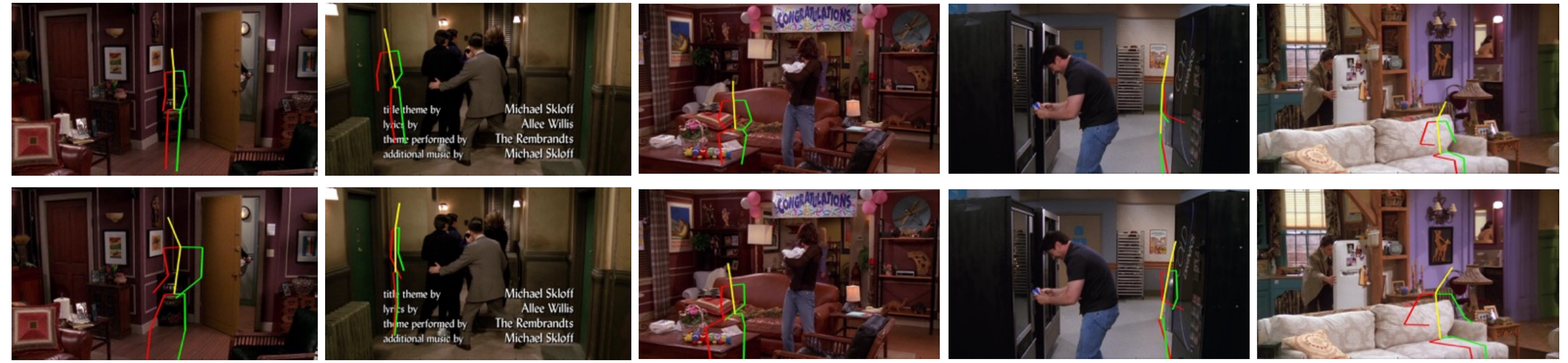}
\vspace{-0.1in}
\caption{\label{fig:results_vae}
Affordance results for VAE with different noise inputs. Each column includes two predictions results for the same inputs. Given the human pose class, we show how VAE can be used to sample multiple scale and deformations.}
\vspace{-0.1in}
\end{figure*}

We have also visualized the results given different noise $z$ as inputs to the VAE Decoder during testing. For the same scene and location, we can generate different poses as 
Fig.~\ref{fig:results_vae}.

\noindent \textbf{Quantitative results.} To show that the generated poses are in reasonable, we first evaluate the performance of our pose classification network. Note that there could be multiple reasonable poses in a given location, thus we show our 30-class classification accuracies given top 1 to top 5 guesses. We test our model on the 3872 samples from ``Friends'', the results is shown in Table~\ref{tbl:result_cls}. We compare models trained on three different sizes of our dataset (2.5K, 10K and 25K). We show that the more data we have, the higher accuracies we can get.
For the heatmap baseline, we use the inner product of the predicted heatmap to assign the output to the cluster centers. We obtain the top-5 assignments and standard evaluation as above for classification performance. As the numbers show, our approach clearly outperforms this heatmap based baseline.

\begin{figure*}[t]
\center
\includegraphics[width=\textwidth]{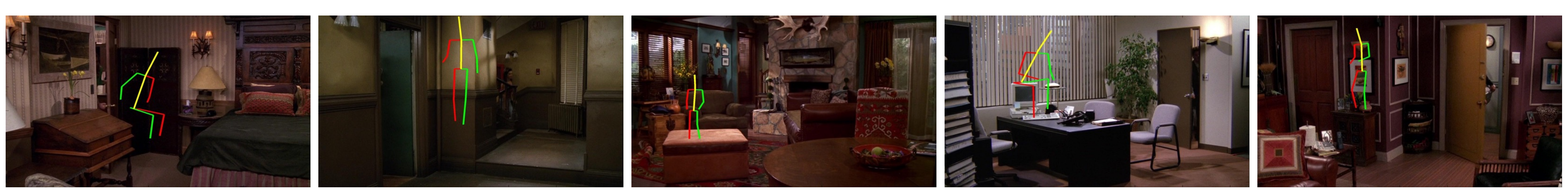}
\vspace{-0.1in}
\caption{\label{fig:neg_data}
Negative samples added in the test dataset. 71\% of the test data is such images and 29\% is positive examples.}
\end{figure*}

\noindent \textbf{Human evaluation.}  We also perform human evaluation on our approach. Given the ground truth pose and predicted pose in the same location of the same image, we ask human which one is more realistic. We find that $46\%$ of the time the turkers think our prediction is more realistic. Note that a random guess is $50\%$, which means our prediction results are almost as real as the ground truth and the turkers can not tell which is generated by our model.

\begin{table} 
\begin{center}
\resizebox{0.5\textwidth}{!}{
\begin{tabular}{|c|c|c|c|c|c|}
  \hline
  Method & Top-1 & Top-2 & Top-3 & Top-4 & Top-5\\ \hline
  HeatMap (Baseline) & 8.4 \% &  19.9\% &  {30.1\%} &  39.1\% &  47.3\% \\
  Training with 2.5K  & 11.7\% &  21.9\% &  {29.7\%} &  36.1\% &  41.8\% \\
  Training with 10K & {13.3\%} &  {23.7\%} & 32.3\%  & {39.7\%} & {46.8\%} \\
  Training with 25K      & \textbf{14.9\%} & \textbf{26.0\%} & \textbf{36.0\%} & \textbf{43.6\%} & \textbf{50.9\%}  \\
  \hline
\end{tabular}
}
\caption{Classification results on the test set.}
\vspace{-0.25in}
\label{tbl:result_cls}
\end{center}
\end{table}

\begin{figure}[t]
\center
\includegraphics[width=0.45\textwidth]{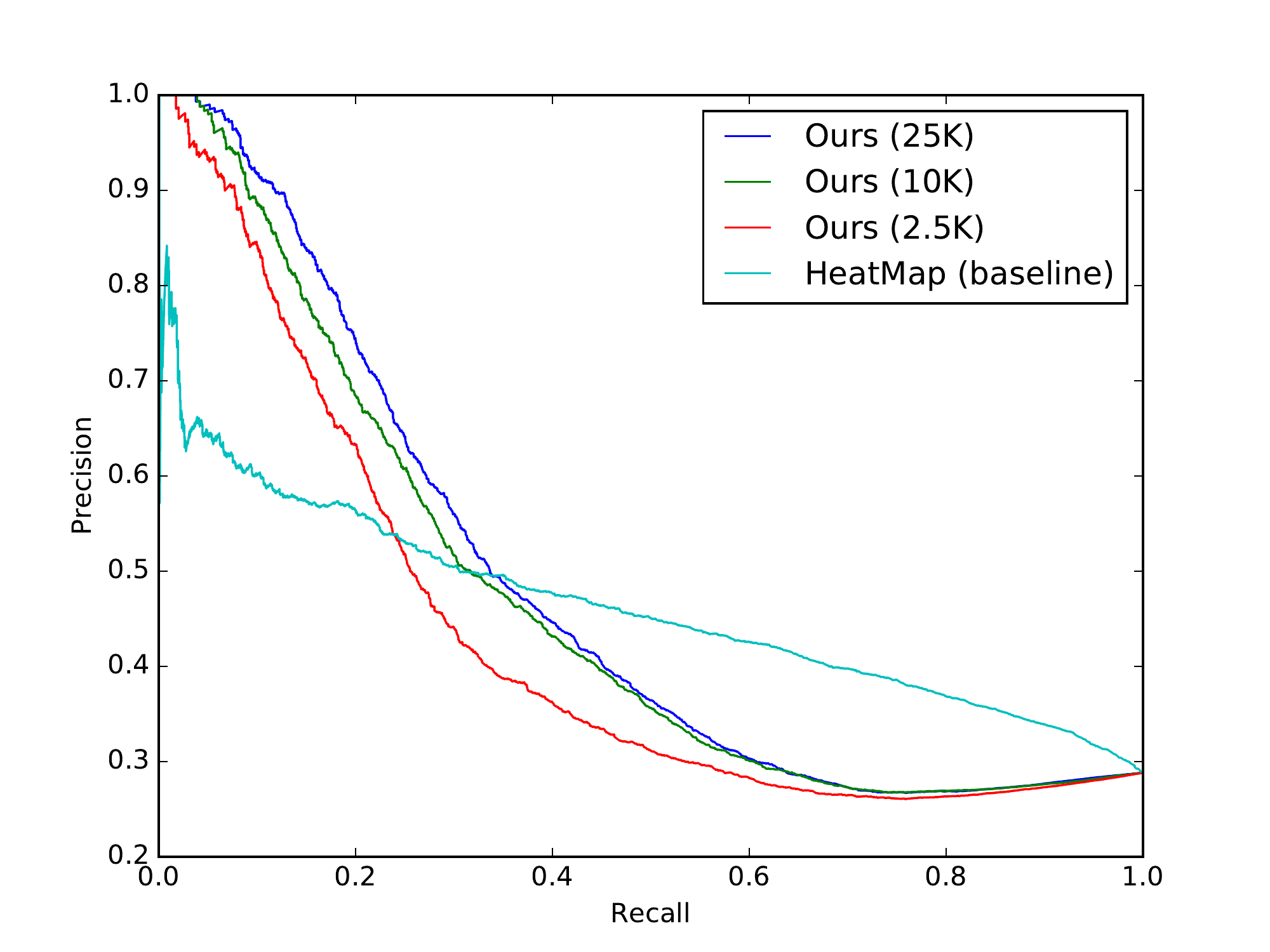}
\vspace{-0.05in}
\caption{\label{fig:pr}
PR curve for our second experiment: given an image and pose, we produce a score on how probable it is. We use these scores to compute the recall-precision graph.}
\vspace{-0.2in}
\end{figure}

\subsection{Classifying given poses in the scenes}
Given a pose in a location of the scene, we want to estimate how likely the pose is using our model. We perform our experiments on 3872 positive samples from ``Friends'' and 9572 negative samples in the same scenes. We show some of the negative samples as Fig.~\ref{fig:neg_data}. Note that although we use negative data in testing, but there is no negative data involved in training. We show the Precision-Recall curve as Fig.~\ref{fig:pr}. Among all of our  approaches, we find that training with 25K data points give best results, which is consistent with the first task.
For the heatmap baseline, we again score each sample as the inner product of 
predicted heatmap with a heatmap representation of the labeled pose. We observe
that the baseline does better than our approach in high-recall regimes, which
can be explained by the fact that training with euclidean loss generates an
averaged-out output, which is less likely to miss a pose.

\vspace{-0.05in}
\section{Conclusion}
\vspace{-0.05in}
In this paper, we present one of the biggest affordance dataset. We use 100 Million frames from seven sitcoms to  extract diverse set of scenes and how actors interact with different objects in those scenes. Our dataset consist of more than 10K scenes and 28K ways humans can interact with these 10K images. We also propose a two step approach to predict affordance pose given an input image and the location. In the first step, we classify which of the 30 pose classes is the likely affordance pose.  Given the pose class and the scene,we then use Variational Autoencoder (VAE) to extract the scale and deformation of the pose.  VAE allows us to sample the distribution of possible poses at test time. Our results indicate that the poses generated by using our approach are quite realistic.

{\footnotesize
\noindent {\bf Acknowledgement}: This work was supported by a research grant from Apple Inc. Any views, opinions, findings, and conclusions or recommendations expressed in this material are those of the author(s) and should not be interpreted as reflecting the views, policies or position, either expressed or implied, of Apple Inc.}

{\small
\bibliographystyle{ieee}
\bibliography{local}
}

\end{document}